\documentclass[journal]{IEEEtran}

\ifCLASSINFOpdf
\else
   \usepackage[dvips]{graphicx}
\fi
\usepackage{url}

\usepackage{graphicx}
\usepackage{hyperref}
\usepackage{amssymb}
\usepackage{amsmath}
\usepackage{multirow}

\usepackage[table]{xcolor}
\definecolor{skyblue}{rgb}{0.88, 0.92, 0.96}

\begin{document}

\title{WaveDH: Wavelet Sub-bands Guided ConvNet \\ for Efficient Image Dehazing}


\author{Seongmin Hwang, Daeyoung Han, 
        Cheolkon Jung, \IEEEmembership{Member, IEEE}, 
        and Moongu Jeon, \IEEEmembership{Member, IEEE}

\thanks{
    S. Hwang is with the Artificial Intelligence Graduate School, Gwangju Institute of Science and Technology (GIST), Gwangju 61005, Republic of Korea (e-mail: sm.hwang@gm.gist.ac.kr).

    C. Jung is with the School of Electronic Engineering, Xidian University, Xi'an 710071, China (e-mail: zhengzk@xidian.edu.cn).

    D. Han, and M. Jeon are with the School of Electrical Engineering and Computer Science, Gwangju Institute of Science and Technology (GIST), Gwangju 61005, Republic of Korea (e-mail: xesta120@gist.ac.kr; mgjeon@gist.ac.kr).

    (Corresponding author: Moongu Jeon).
    }}

\markboth{Journal of \LaTeX\ Class Files, Vol. 14, No. 8, August 2015}
{Shell \MakeLowercase{\textit{et al.}}: Bare Demo of IEEEtran.cls for IEEE Journals}
\maketitle

\begin{abstract}
The surge in interest regarding image dehazing has led to notable advancements in deep learning-based single image dehazing approaches, exhibiting impressive performance in recent studies. Despite these strides, many existing methods fall short in meeting the efficiency demands of practical applications. In this paper, we introduce WaveDH, a novel and compact ConvNet designed to address this efficiency gap in image dehazing. Our WaveDH leverages wavelet sub-bands for guided up-and-downsampling and frequency-aware feature refinement. The key idea lies in utilizing wavelet decomposition to extract low-and-high frequency components from feature levels, allowing for faster processing while upholding high-quality reconstruction. The downsampling block employs a novel squeeze-and-attention scheme to optimize the feature downsampling process in a structurally compact manner through wavelet domain learning, preserving discriminative features while discarding noise components. In our upsampling block, we introduce a dual-upsample and fusion mechanism to enhance high-frequency component awareness, aiding in the reconstruction of high-frequency details. Departing from conventional dehazing methods that treat low-and-high frequency components equally, our feature refinement block strategically processes features with a frequency-aware approach. By employing a coarse-to-fine methodology, it not only refines the details at frequency levels but also significantly optimizes computational costs. The refinement is performed in a maximum 8$\times$ downsampled feature space, striking a favorable efficiency-vs-accuracy trade-off. Extensive experiments demonstrate that our method, WaveDH, outperforms many state-of-the-art methods on several image dehazing benchmarks with significantly reduced computational costs. Our code is available at \href{https://github.com/AwesomeHwang/WaveDH}{https://github.com/AwesomeHwang/WaveDH}.
\end{abstract}

\begin{IEEEkeywords}
Single image dehazing, deep learning, wavelet sub-bands, frequency awareness
\end{IEEEkeywords}

\IEEEpeerreviewmaketitle

\section{Introduction}

\IEEEPARstart{H}{aze}, a natural atmospheric phenomenon, induces visible degradation in visual quality by affecting object appearance and contrast through color and texture distortion. Images captured in hazy conditions pose challenges for subsequent tasks such as object detection \cite{sindagi2020prior}, vehicle re-identification \cite{chen2022sjdl}, and scene understanding \cite{sakaridis2018semantic}. Consequently, the removal of haze from images is a critical concern in low-level vision, essential for developing effective computer vision systems. Image dehazing aims to restore the latent haze-free scene from its hazy observation, presenting an inherently ill-posed and challenging problem. For the single image dehazing task, there is a widely used atmospheric scattering model \cite{ASM} which estimates the clear image from a single hazy input, expressed as:
\begin{equation}
I=J(x)t(x)+A(1-t(x)),
\label{eqn1:asm}
\end{equation}
where $I$ is the captured hazy image, $J$ is the corresponding clear scene, $A$ is the global atmospheric light, and $t$ is the medium transmission, which is formulated by the scene depth $d$ with the atmosphere scattering parameter $\beta$ as:
\begin{equation}
t(x)=e^{-\beta d(x)}
\end{equation}

 \begin{figure}[t]
 \centering
    \includegraphics[scale=0.35]{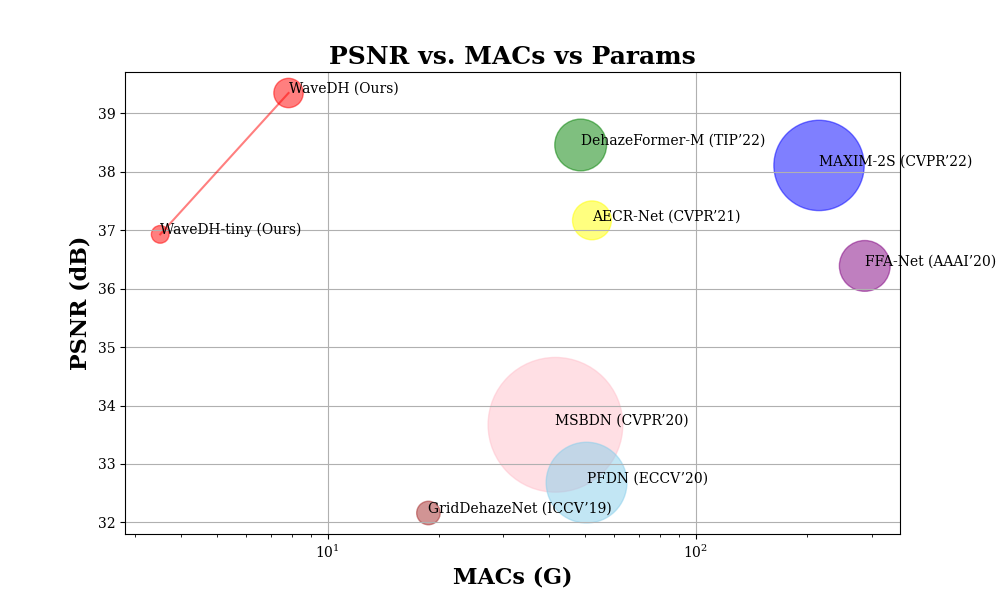}
    \caption{Comparison of WaveDH with other image dehazing methods on the SOTS indoor set \cite{RESIDE}. The circle size is proportional to the number of model parameters. Note that our WaveDH achieves superior PSNR and also maintains lower model complexity.
    }
    \label{fig0:psnr_vs_flops}
        \vspace{-0.5cm}
 \end{figure}

In recent years, deep learning-based methods, leveraging the powerful learning capabilities of convolutional neural networks (CNNs), have shown outstanding performance in various computer vision tasks including single image dehazing. While some methods \cite{dehazenet,MSCNN,DCPDN,DehazeRDN,RefineDNet} still adhere to the atmospheric scattering model, recent studies \cite{Griddehazenet, EPDN, GCANet, Hardgan, MSBDN, ffanet, aecr, Msaff} prefer an end-to-end approach, achieving superior results by predicting the latent haze-free image or its residuals versus the hazy image. Very recently, vision Transformer (ViT) \cite{ViT} has also shown promise in image dehazing \cite{dehazeformer, dehamer, MB_TaylorFormer} based on its strong capability to model long-range dependencies. Despite the remarkable advancements achieved by these networks, they often rely on stacking deeper and more complex models, posing challenges for deployment on resource-limited devices such as surveillance cameras and mobile phones in real-world scenarios. This motivates the need for designing fast and lightweight deep models that offer a better trade-off between performance and computational complexity.

To address the challenges posed by heavy deep dehazing models, several methods have been proposed in recent years. Zhang et al. \cite{famednet} adopted AOD-Net's formulation and proposed a fast and accurate multi-scale dehazing network (FAMED-Net), which comprises encoders at three scales and a fusion module. Wu et al. \cite{aecr} adopted autoencoder-like (AE) framework to make dense convolution computation in the low-resolution space and also reduce the number of layers to design compact model. Another approach is LD-Net \cite{ldnet} which jointly estimates both the transmission map and the atmospheric light, contributing to model efficiency. Although these methods offer compact architectures improving efficiency, the trade-off between efficiency and dehazing performance remains sub-optimal. We believe there is still large room to achieve a better trade-off.

In the pursuit of mitigating computation costs, two primary options are typically considered. The first involves the use of manually designed lightweight structures \cite{mobilenet, mobilev2, shufflev2, efficientv2}, which is a strategy that has been proposed over the past few years. Among these structures, depthwise separable convolutions (DSConv) \cite{mobilenet} stand out as one of the most fundamental architectures. On the other hand, aside from employing lightweight structures, computation costs can be alleviated by reducing the feature size \cite{mobilev3, ghostnet}. While downsampling operations, such as max pooling in \cite{famednet}, can effectively decrease computational costs, these pooling-based operations are prone to information dropping, particularly high-frequency components crucial for texture details. Consequently, this can adversely impact the overall reconstruction quality. Recent studies (\textit{e.g.}, \cite{shift}) further emphasize that the applying pooling operations in CNNs could hurt the shift-equivariance of deep networks.

In this paper, our aim is to design an efficient and accurate dehazing network, and to that end, we propose WaveDH—a novel wavelet sub-bands guided dehazing ConvNet. Our WaveDH improves network effectiveness by optimizing up-and-downsampling processes through non-aggressive down-sampling. Additionally, we enhance network efficiency based on a frequency-aware feature refinement block which is designed for efficient representation learning. As the name suggests, the WaveDH is built upon wavelet transform, which decomposes input into four sub-bands for low-and-high frequency components. Since our upsampling block takes the high-frequency components returned from the same level downsampling block as an additional input, the up-and-downsampling processes provide multi-scale information while preserving high-frequency details. 

Furthermore, our frequency-aware feature refinement block plays a pivotal role in elevating representation learning efficiency. Operating in a coarse-to-fine manner, this block refines both low-and-high frequency information, ensuring that WaveDH captures intricate scene details comprehensively. Within our refinement block, the Feature Mixing Block (FMB) focuses on learning structure and context at a coarse-grain level with the low-frequency sub-bands. The high-frequency components are refined through a feature distillation mechanism. This mechanism efficiently interacts with the refined low-frequency and high-frequency information at a fine-grained level, ensuring that our WaveDH produces dehazed images with enhanced clarity and fidelity. By leveraging the invertible properties of the wavelet transform, we enhance low-frequency information in a downsampled feature space by a factor of two at each level, leading to a superior trade-off between performance and computational efficiency. Fig. \ref{fig0:psnr_vs_flops} shows
the comparison of our WaveDH with other state-of-the-art image dehazing methods on the SOTS indoor set.

Our key contributions can be summarized as follows:
\begin{itemize}
    \item We propose a wavelet sub-bands guided ConvNet, abbreviated as WaveDH, for fast and accurate single image dehazing. Taking advantage of discrete wavelet transform, our WaveDH achieves a superior trade-off between efficiency and performance.
    
    \item We design wavelet-guided up-and-downsampling blocks that utilize inherent lossless and invertible downsampling properties of the wavelet transform for optimized upsampling and downsampling.
  
    \item We present a frequency-aware feature refinement block to efficiently learn intermediate feature representations. Our refinement processing adaptively handles features in a coarse-to-fine manner based on frequency-awareness for computational efficiency.
\end{itemize}

\section{Related Work}
\subsubsection{Single Image Dehazing}

Single image dehazing is a challenging task due to the lack of information in hazy conditions. Traditional methods mainly rely on an atmospheric scattering model \cite{ASM} and the handcrafted priors. Notable among these is the Dark Channel Prior (DCP) \cite{DCP}, which estimates the medium transmission map. Subsequently, various priors-based methods have been proposed, including color attenuation prior (CAP) \cite{CAP} and non-local prior \cite{NLP}. However, these prior-based methods may lead to unrealistic results in complex scenes that do not conform to these priors. 

In recent years, learning-based methods using large-scale datasets have dominated single image dehazing. Pioneering works \cite{dehazenet, MSCNN} employed convolutional neural networks (CNNs) to estimate the transmission map and global atmospheric light in the physics model to restore a latent hazy-free scene. Since the advent of the pioneer works, deep learning-based approaches are explored to achieve more accurate results. Li {\em et al.} \cite{aodnet} reformulated atmospheric model (Eq. (\ref{eqn1:asm})) and proposed an all-in-one dehazing network (AOD-Net). Zhang {\em et al.} \cite{DCPDN} proposed a densely connected pyramid network (DCPDN) that uses two sub-networks to estimate the transmission map and global atmospheric light, respectively. On the other hand, Liu et al. \cite{Griddehazenet} proposed an attention-based multi-scale network (GridDehazeNet), which learns the feature map to restore the hazy-free image directly instead of estimating the transmission map. FFANet \cite{ffanet} proposed a deep network that introduces feature attention (FA) blocks that leverage both channel and pixel attention to improve haze removal. 

Since Dosovitskiy et al. \cite{ViT} introduced Transformer to computer vision, the Vision Transformer (ViT) architectures have demonstrated the capability of replacing CNNs. Song et al. \cite{dehazeformer} proposed the Dehazeformer which can be considered as a combination of Swin Transformer and U-Net and showed superior performance. DeHamer \cite{dehamer} which is a hybrid model of CNN and Transformer for image dehazing, which can aggregate global attention and local attention. 

While significant progress has been made by aforementioned methods, the reliance on deeper and more complex models for performance improvement hinders real-world deployment. Additionally, most existing methods are spatial-domain-centric, neglecting the exploitation of frequency domain information to estimate clear scene.

 \begin{figure*}[t]
 \centering
    \includegraphics[scale=0.52]{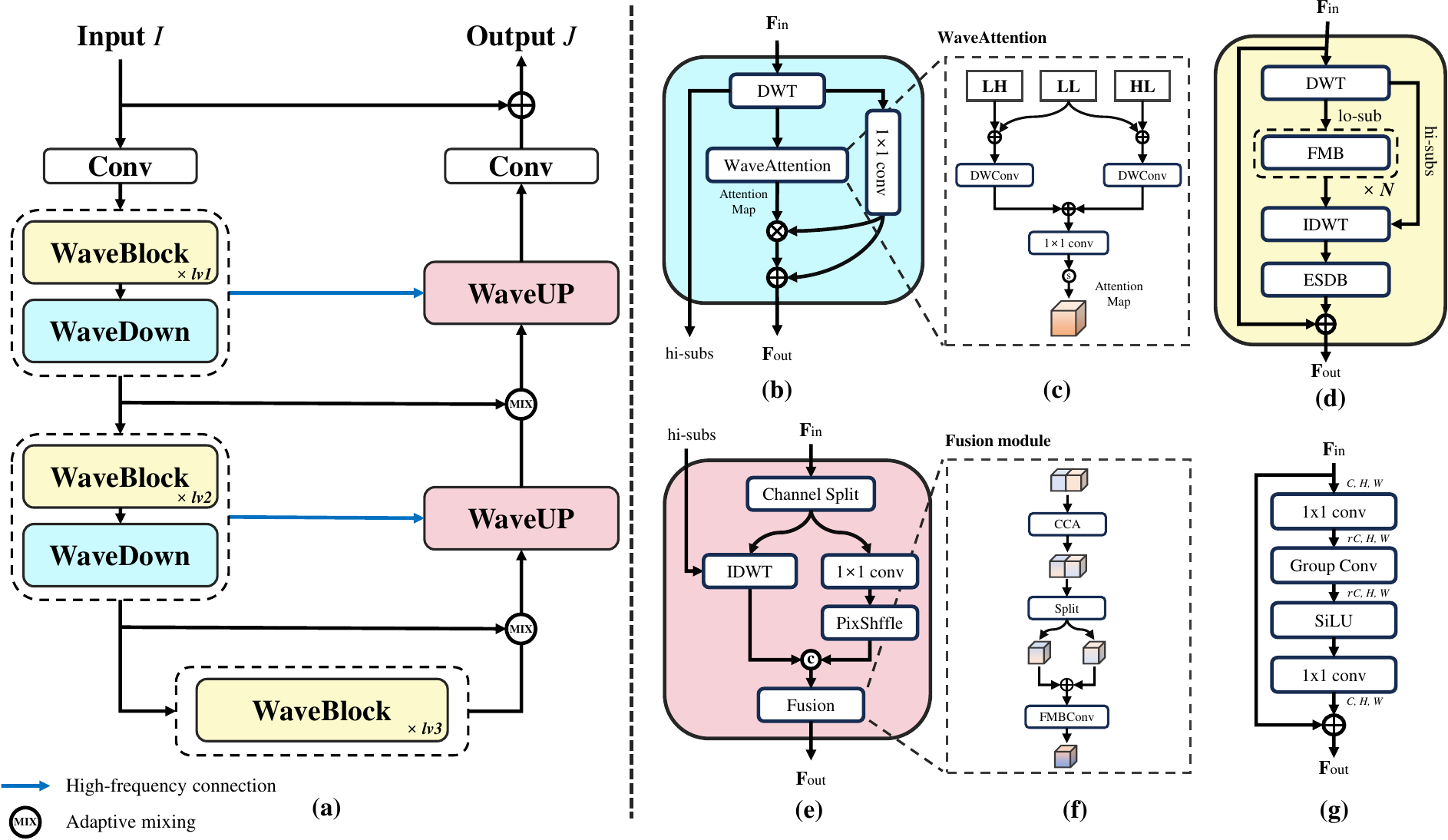}
    \caption{Overview of WaveDH architecture. (a) Depicts the overall architecture consists of various modules and their contribution to efficient image dehazing. (b) Shows our downsampling block using WaveAttention. (c) Illustrates the WaveAttention module for noise suppression and detail preservation. (d) Presents the WaveBlock for frequency-aware feature learning. (e) Represents the WaveUp block based on dual-upsample and fusion mechanism. (f) Represents the fusion module crucial for interacting dual-upsampled features. (g) Presents the modified Fused-MBConv (FMBConv) employing group convolution to optimize parameter usage and efficiency.}
    \label{fig0:overall}
        \vspace{-0.5cm}
 \end{figure*}

\subsubsection{Wavelet-based Approaches in Computer Vision}

Wavelet decomposition, which widely used in signal processing \cite{singularity, noisedata}, separates low-and-high frequency components from signals. Its application in deep learning architectures, such as CNNs and transformers, enhances spatial and frequency information, improving their performance in various vision tasks. Some works have employed wavelet transform to network design to enhance visual representation learning \cite{WaveCNN, WaveVIT}. Furthermore, it has been extended to diverse tasks such as style transfer \cite{style}, face hallucination \cite{facehallucination} and image generation \cite{WaveGAN, Styleswin}.

Given its ability to decompose the input image into multi-scale sub-images and its invertibility, wavelet transform finds extensive applications in low-level vision tasks, particularly in image restoration. Bae et al. \cite{waveIR} proposed that deep residual learning in the CNN feature space over wavelet sub-bands can be beneficial for image restoration. Liu et al. \cite{MWCNN} introduced a multi-level wavelet CNN (MWCNN) where wavelet transform is employed to reduce the size of feature maps, thereby improving the trade-off between image restoration performance and efficiency. Guo et al. \cite{DWSR} proposed DWSR, a method that takes low-resolution wavelet sub-bands and outputs residuals of corresponding sub-bands of high-resolution wavelet coefficients to recover missing details for image super-resolution. DeWRNet \cite{DeWRNet} presented an image super-resolution enhancement technique that trains low-and-high frequency sub-images with different models, utilizing high-frequency sub-images and input images derived from stationary wavelet transform for interpolation. Recent studies on deraining \cite{deraining1, SWAL} have found that wavelet transform is beneficial in decomposing rain image features into different scale information while preserving all information.

When it comes to image dehazing, the idea of exploiting  a wavelet decomposition is not new. In \cite{neurodehaze} and \cite{accessdehaze}, the feature learning is mainly achieved in the wavelet domain to obtain high-quality haze-free output. Specifically, Khan et al. \cite{neurodehaze} introduced a hybrid approach to dehaze the corrupted image by decomposing high-frequency sub-bands and approximating the low-frequency sub-band of the given hazy image using the wavelet domain. In \cite{accessdehaze}, a multi-scale wavelet and non-local dehazing method were introduced, where non-local dehazing and wavelet denoising are respectively carried out on the low-and-high frequency sub-images to remove haze and noise. Despite the remarkable progress achieved by these approaches, existing methods do not fully exploit the sub-bands domain information and invertible properties of wavelet transform, limiting their performance. 

In this paper, our main focus is on exploring the feasibility of a wavelet-guided ConvNet for efficient image dehazing to facilitate practical applications. In contrast to existing methods, our feature refinement block efficiently refines features in a coarse-to-fine fashion by separating low-frequency and high-frequency features. Moreover, by leveraging the invertible properties of wavelet transform, we refine low-frequency information $\times 2$ downsample feature space at each level without loss of information leading to a better trade-off between performance and computational efficiency.

\section{Proposed Method}
In this section, we present our proposed single image dehazing network, WaveDH. The method is designed to be efficient and lightweight, utilizing a hierarchical architecture with wavelet-guided up-and-downsampling blocks, and frequency-aware feature refinement blocks. We begin with an overview of the overall architecture, followed by a detailed presentation of the wavelet-guided up-and-downsampling blocks. Finally, we introduce frequency-aware feature refinement block.

\subsection{Overall Architecture}
Fig. \ref{fig0:overall} illustrates the comprehensive pipeline of our approach, WaveDH. Our WaveDH has a highly hierarchical architecture (\textit{i.e.}, U-Net \cite{unet} like architecture), complemented by skip connections.

Given a hazy image $I \in \mathbb{R}^{H \times W \times C}$, where $H \times W$ represents spatial resolution and $C$ is the number of channels, the WaveDH initiates by extracting low-level features $Z_0 \in \mathbb{R}^{H \times W \times D}$ through a convolutional operation with $D$ channels. These low-level features traverse a symmetric hierarchical encoder-decoder structure, culminating in high-level deep features $Z_5 \in \mathbb{R}^{H \times W \times 2D}$. Departing from conventional pooling-based downsamplers and popularly-used upsamplers like pixelshuffle \cite{pixshuffle} and transposed convolution \cite{transposed}, we introduce wavelet-based down-and-upsampling blocks (\textit{i.e.}, WaveUP and WaveDown). At the core of the model, frequency-aware bottleneck blocks (\textit{i.e.}, WaveBlock) are strategically crafted to adeptly handle both low-and-high frequency information. The WaveBlocks are selectively employed in the encoder stage and at the base of the model. In the last step, a convolutional layer is employed to map the features $Z_5$ to the residual image $R \in \mathbb{R}^{H \times W \times C}$, and the final hazy-free estimation $J$ is yielded as $J = I + R$. In the following subsections, we delve into the details of the newly introduced components.

\subsection{Wavelet-guided downsampling and upsampling blocks}
Commonly adopted downsampling operations, such as max pooling and average pooling, inevitably lead to information loss. In addition to these popular pooling-based methods, a strided convolution, considered as a learnable pooling operation, offers a promising alternative that enhances the expressive capabilities of the network. However, it comes with an increase in the number of trainable parameters and lacks invertibility. To overcome these challenges, we introduce wavelet-sub bands guided upsampling and downsampling blocks, leveraging the inherent properties of wavelet transform to perform lossless and invertible downsampling. Fig. \ref{fig0:overall} (b) and Fig. \ref{fig0:overall} (e) depict the structure of our blocks in detail.

\subsubsection{Downsampling block}
Our downsampling block, WaveDown, first applies discrete wavelet transform to the given 2D feature map $F \in \mathbb{R}^{H \times W \times D}$, decomposing it into four wavelet sub-bands. Specifically, DWT uses the low-pass filter $L=\left[\begin{array}{ll}1 / \sqrt{2} & 1 / \sqrt{2}\end{array}\right]$ and high pass filter $H=\left[\begin{array}{ll}-1 / \sqrt{2} & 1 / \sqrt{2}\end{array}\right]$ to construct four kernels with stride 2 (\textit{i.e.}, $LL^T, LH^T, HL^T$, and $HH^T$). Next, these kernels decompose the input into the four wavelet sub-bands: $F_{ll} \in \mathbb{R}^{\frac{H}{2} \times \frac{W}{2} \times D}, F_{lh} \in \mathbb{R}^{\frac{H}{2} \times \frac{W}{2} \times D}, F_{hl} \in \mathbb{R}^{\frac{H}{2} \times \frac{W}{2} \times D}$, and $F_{hh} \in\mathbb{R}^{\frac{H}{2} \times \frac{W}{2} \times D}$. $F_{ll}$ is a low-frequency approximation that retains the main structural information of the feature map at a coarse-grained level. $F_{lh}, F_{hl}$, and $F_{hh}$ are high-frequency components that provide detailed information at a fine-grained level while retaining significant amount of noise of the feature map.

Rather than discarding all high-frequency components as noise, as done in \cite{Wavecnet, WavePool}, which contains a rich amount of detailed information, we utilize not only low-frequency component but also high-frequency components. To reduce noise interference, we introduce a squeeze-and-attention mechanism. Specifically, we concatenate the four wavelet sub-bands along the channel dimension to form $\hat{F} = [F_{ll}, F_{lh}, F_{hl}, F_{hh}] \in \mathbb{R}^{H \times W \times 4D}$. We then squeeze all information into a low-dimensional manifold across channels using a $1\times1$ convolution layer, mapping the input into a compressed space, expressed as:
\begin{equation}
\tilde{F} = \mathbf{W}_{p_1}(\hat{F}),
\end{equation}
where $\mathbf{W}_{p_1} \in \mathbb{R}^{4D \times 2D}$ is the squeeze operation, \textit{i.e.}, point-wise convolution, reducing the input channels. $\tilde{F}$ contains compressed information from $\hat{F}$ (\textit{e.g.}, high- and low-frequency information and noise). 

To generate an attention map, $F_{ll}$ and $F_{lh}$, as well as $F_{ll}$ and $F_{hl}$, are elementwise added. These sums then pass through 3$\times$3 depthwise convolution layers, respectively, and the results are elementwise added again. Finally, the attention map is generated by sigmoid function. The entire process is illustrated in Fig. \ref{fig0:overall} (c) and can be formulated as follows:
\begin{equation}
M = \sigma(\mathbf{W}_{p_2}(\mathbf{W}_{d_1}(F_{ll}+F_{lh}) + \mathbf{W}_{d_2}(F_{ll}+F_{hl}))),
\end{equation}
where $\sigma$ refers to sigmoid function, $\mathbf{W}_{p_2} \in \mathbb{R}^{D \times 2D}$ is a point-wise convolution operation increasing the input channels, and $\mathbf{W}_{d_1}$ and $\mathbf{W}_{d_2}$ are depthwise convolution operations. Note that, as high-frequency component $F_{hh}$ contains excessive noise information, it is discarded to generate a more convincing attention map.

\begin{figure}[t]
 \centering
    \includegraphics[scale=0.38]{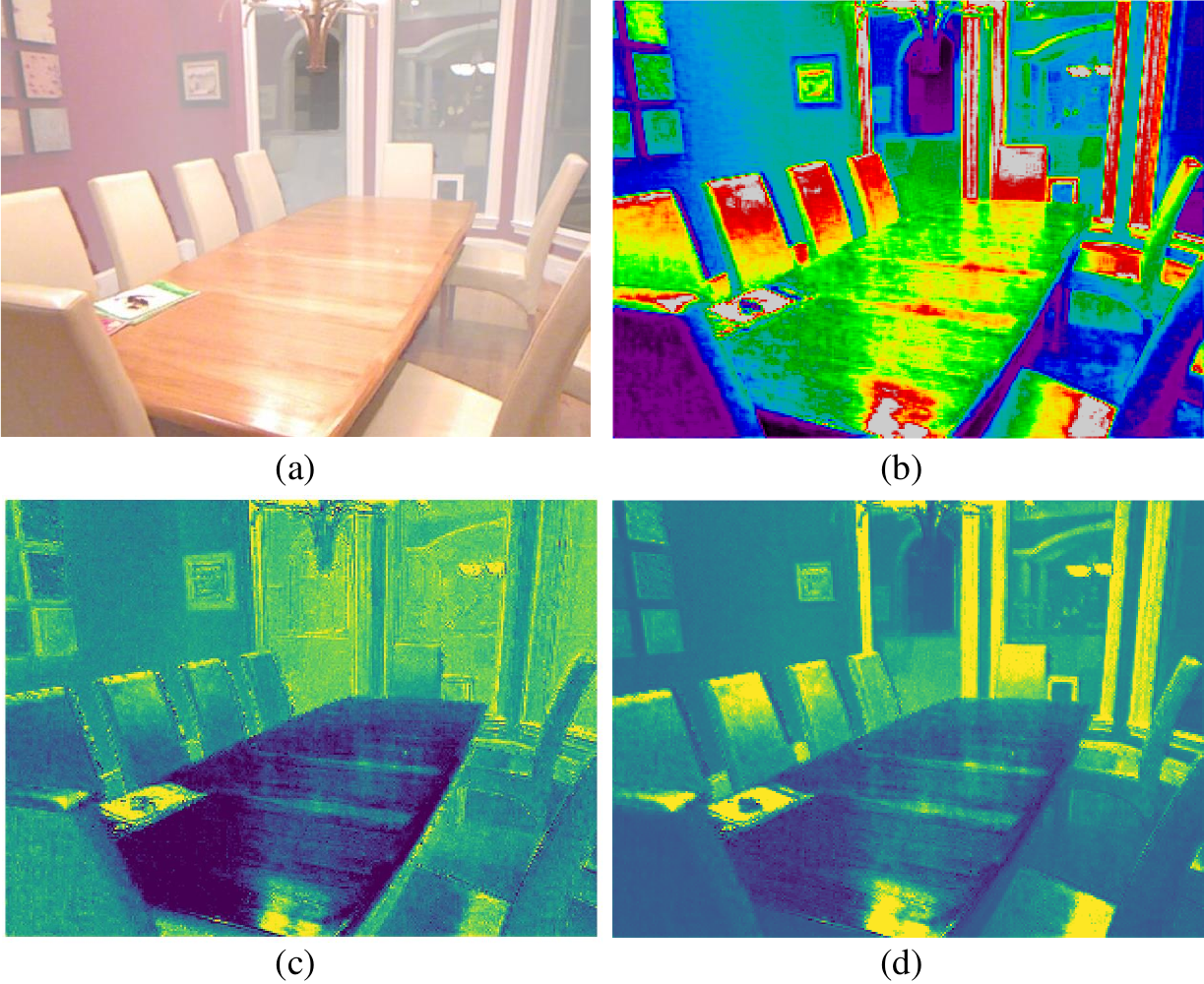}
    \caption{This figure depicts the enhanced feature map via the wavelet attention mechanism. (a) Original hazy input. (b) Visualization of attention map. (c) Visualization of feature map before WaveAttention. (d) Visualization of feature map after WaveAttention. The WaveAttention refines the feature map by suppressing noise and emphasizing informative details, contributing to improved dehazing performance.}
    \label{fig2:wave_att}
        \vspace{-0.5cm}
 \end{figure}

The generated attention map $M$ is then Hadamard multiplied by $\tilde{F}$ to suppress noise while boosting useful coarse- and fine-grained information (\textit{e.g.}, low-frequency context and high-frequency details) as illustrated in Fig. \ref{fig2:wave_att}, this step is formulated as:
\begin{equation}
\tilde{F}_{att} = \tilde{F} \odot M,
\end{equation}
where $\odot$ denotes Hadamard multiplication.

Finally, by elementwise adding $\tilde{F}_{att}$ and $\tilde{F}$, we obtain the final downsampled output feature maps:
\begin{equation}
Y = \tilde{F}_{att} + \tilde{F},
\end{equation}
where $Y \in \mathbb{R}^{\frac{H}{2} \times \frac{W}{2} \times 2D}$ is the final output. It is worth mentioning that our downsampling blocks additionally return high-frequency components $\hat{F}_{high} = [F_{lh}, F_{hl}, F_{hh}] \in \mathbb{R}^{H \times W \times \frac{3}{2}D}$ to provide the upsampling blocks with frequency cues.

\subsubsection{Upsampling block}
To fully exploit the invertible property of the wavelet transform, we introduce a dual-upsample and fusion mechanism, as illustrated in Fig. \ref{fig0:overall} (e). The wavelet-guided upsampling block, WaveUP, consists of two modules: Inverse Discrete Wavelet Transform (IDWT) and PixelShuffle, along with a novel fusion module. PixelShuffle is more efficient than transposed convolution to enlarge the spatial resolution alleviating checkerboard artifacts associated with transposed convolution. However, relying solely on PixelShuffle may lead to a lack of high-frequency details in lightweight models due to its limited representation capacity. To address this, our upsampling block takes high-frequency components returned from the corresponding downsampling block at the same level as an additional input.

Starting with the input feature $F \in \mathbb{R}^{H \times W \times D}$, we split it into $F_1 \in \mathbb{R}^{H \times W \times \frac{D}{2}}$ and $F_2 \in \mathbb{R}^{H \times W \times \frac{D}{2}}$ along the channel dimension:
\begin{equation}
F_1, F_2 = \text{split}(F).
\end{equation}

To save the number of parameters in the upsampling block, we opt for a $1 \times 1$ convolutional layer to expand the channel of feature map $F_1$, followed by passing it through the PixelShuffle layer. Simultaneously, $F_2$ is concatenated with high-frequency components $\hat{F}{high}$ from the corresponding downsampling block. This concatenated input is fed into the IDWT layer:
\begin{align}
& \hat{F_1} = \mathrm{PixShuffle}(\mathbf{W}_{p_3}(F_1)), \\
& \hat{F_2} = \mathrm{IDWT}(Concat(F_2, \hat{F}_{high})),
\end{align}
where $\mathbf{W}{p_3} \in \mathbb{R}^{\frac{D}{2} \times 2D}$ is a $1 \times 1$ convolution expanding the channel dimension, and $\mathrm{PixShuffle}(\cdot)$ indicates the PixelShuffle layer. $Concat(\cdot)$ denotes the concatenation operation along the channel dimension. The two outputs, $\hat{F_1} \in \mathbb{R}^{2H \times 2W \times \frac{D}{2}}$ and $\hat{F_2} \in \mathbb{R}^{2H \times 2W \times \frac{D}{2}}$, are then fused using our proposed fusion module (See Fig. \ref{fig0:overall} (f)).

 \begin{figure}[t]
 \centering
    \includegraphics[scale=0.5]{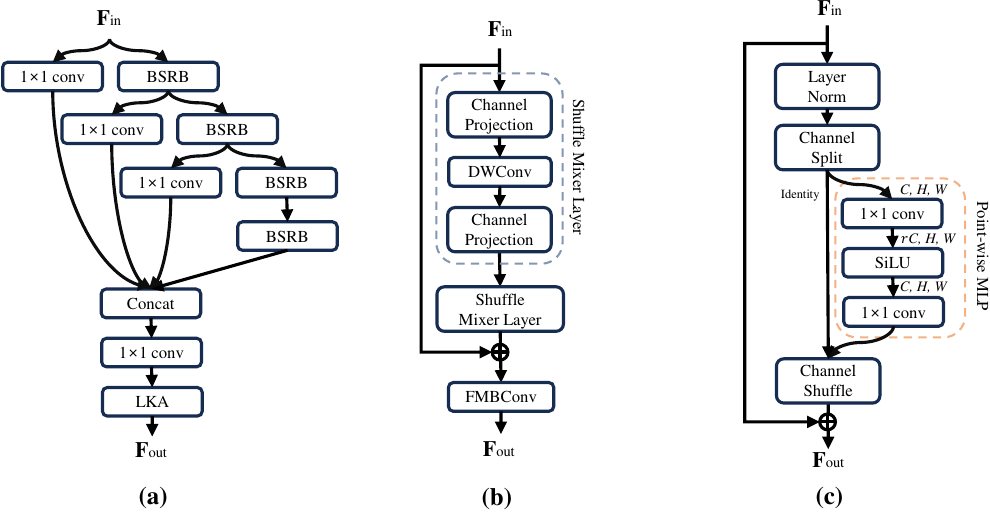}
    \caption{Overview of the proposed WaveBlock components. (a) The architecture of Efficient Separable Distillation Block (ESDB), employing a series of 1x1 convolutional layers with Blueprint Shallow Residual Blocks (BSRBs) for feature distillation. (b) The architecture of Feature Mixing Block (FMB), consisting of depthwise convolution (DWConv), a shuffle mixer layer, and two (c) channel projection modules.}
    \label{fig1:blocks} 
        \vspace{-0.5cm}
 \end{figure}

To selectively search their useful information, our feature fusion process employs co-attention for efficient feature interaction. As depicted in Fig. \ref{fig0:overall} (f)), our fusion module consists of contrast-aware attention \cite{cca} (CCA) and a Fused-MBConv \cite{efficientv2} (FMBConv) with some modification. We concatenate $\hat{F_1}$ and $\hat{F_2}$ along the channel dimension, which then pass through a CCA layer playing a pivotal role in feature interaction. The interaction of the dual-upsampled features produces channel-wise co-attention scores, used for recalibrating the concatenated features. This process is expressed as:
\begin{equation}
\tilde{F} = \mathrm{CCA}(Concat(\hat{F_1}, \hat{F_2})),
\end{equation}
where $\tilde{F} \in \mathbb{R}^{2H \times 2W \times D}$ is the resultant feature map, and $\mathrm{CCA}(\cdot)$ indicates the CCA layer. Subsequently, we split the feature map $\tilde{F}$ into two along the channel dimension and elementwise add each divided feature map:
\begin{align}
& \tilde{F_1}, \tilde{F_2} = split(\tilde{F}), \\
& \hat{R} = \tilde{F_1} + \tilde{F_2},
\end{align}
where $\hat{R} \in \mathbb{R}^{2H \times 2W \times \frac{D}{2}}$ is the roughly fused result. This process can be interpreted as a channel-wise weighted summation using co-attention weights. Finally, the output is obtained by applying an FMBConv layer to further refine $\hat{R}$:
\begin{equation}
R = H_{F}(\hat{R}),
\end{equation}
where $R$ is the final fusion result, and $H_{F}$ refers to FMBConv.

The original FMBConv significantly increases the number of parameters and flops. Therefore, \cite{shufflemixer} removed the SE layer and limited the hidden dimension expansion to mitigate this problem. Believing that there is room for a better trade-off between computational cost and performance, we further improved it, as depicted in Fig. \ref{fig0:overall} (g). Group convolution, known for significantly reducing parameters compared to standard convolution, is employed in our block. We experimentally set the parameter $r_{conv}$ as $\frac{3}{2}$. The proposed feature fusion process for dual-upsampling results further preserves the consistent detailed texture structures of the high-frequency.

\subsection{Frequency-aware feature refinement block} 
Our feature refinement block, named WaveBlock, is designed with the goal of frequency-aware discriminative feature learning. As depicted in Fig. \ref{fig0:overall} (d), this block consists of four key components: Feature Mixing Block (FMB) \cite{shufflemixer}, Efficient Separable Distillation Block (ESDB) \cite{esdb}, and Discrete Wavelet Transform (DWT) and Inverse DWT (IDWT) layers.

In most dehazing methods, there is often a tendency to equally treat both the low-and-high frequency components of the feature map, overlooking significant distinctions between different frequency. This approach not only poses challenges in terms of reducing computational costs and achieving effective dehazing but also lacks flexibility in handling diverse types of information. Typically, low-frequency components convey smoother appearances in global regions, whereas high-frequency components capture local regions with richer details such as edges, textures, and other intricate features.

Our approach starts by employing wavelet decomposition to split the feature map $F \in \mathbb{R}^{H \times W \times D}$ into a low-frequency sub-band $F_{ll} \in \mathbb{R}^{\frac{H}{2} \times \frac{W}{2} \times D}$ and the concatenation of high-frequency sub-bands $F_{high} \in \mathbb{R}^{\frac{H}{2} \times \frac{W}{2} \times 3D}$. Subsequently, the low-frequency approximation $F_{ll}$ passes through further processing within the FMB(s):
\begin{equation}
\hat{F_{ll}} = H_{FMBs}(F_{ll}),
\end{equation}
where $H_{FMBs}$ refers to FMB(s), $\hat{F_{ll}}$ is the enhanced low-frequency component. Note that compared with the original FMB, we change the channel expansion ratio $r_{mlp}$ as $\frac{5}{4}$ (Fig. \ref{fig1:blocks} (c)) from 2 for parameter and computation efficiency. The reason we employ FMB to enhance $F_{ll}$ is that FMB excels in extracting global features through depth-wise convolutions with large kernel sizes, thereby maintaining a parameter and computation-efficient design. This enhances the low-frequency features, containing main structural information at a coarse-grained level. Importantly, $\hat{F_{ll}}$ has reduced spatial resolution, leading to substantial savings in computational costs. It's noteworthy that this reduction in spatial resolution contributes to a more efficient computation process, making our approach particularly advantageous for resource-limited scenarios.

Next, $\hat{F_{ll}}$ is concatenated with the high-frequency sub-bands $F_{high}$, and the Inverse DWT (IDWT) is applied to reconstruct the high-resolution feature map:
\begin{equation}
\hat{F} = \mathrm{IDWT}(Concat(\hat{F_{ll}}, F_{high})),
\end{equation}
where $\hat{F} \in \mathbb{R}^{H \times W \times D}$ is the resultant feature map, now enhanced with low-frequency sub-band information. However, the high-frequency components containing rich detailed information (\textit{e.g.}, object texture details) also need to refine features at a fine-grained level.

For the effective enhancement of high-frequency details, we employ a feature distillation mechanism \cite{idn, imdn, esdb}. In this process, the feature map $\hat{F}$ is fed into ESDB for feature distillation, where it efficiently interacts with refined low-frequency and high-frequency information at a fine-grained level to boost high-frequency information. This pipeline can be expressed as:
\begin{equation}
\tilde{F} = H_{ESDB}(\hat{F}),
\end{equation}
where $H_{ESDB}$ refers to ESDB, and $\tilde{F}$ is the final result. The ESDB replaces standard convolution with a well-designed depthwise separable convolution (\textit{i.e.}, blueprint separable convolution.) to save computation, enabling efficient feature extraction at a fine-grained level. We remove the Contrast-aware Attention (CCA) and Enhanced Spatial Attention (ESA) \cite{esa} modules in ESDB, and replace them with the Large Kernel Attention (LKA) module \cite{lka}, to further enhance computational efficiency. Additionally, the residual connection in ESDB is removed in favor of our residual learning approach (See Fig. \ref{fig0:overall} (d) and Fig. \ref{fig1:blocks} (a)).

Our feature refinement block first handles the main structural information at a coarse-grained level and then deals with detailed information at a fine-grained level. Therefore, the feature processing proceeds in a coarse-to-fine fashion.

\begin{table}[t]
\centering
\caption{The detailed architecture configuration.}
\scalebox{1.0}{
\begin{tabular}{l|ccc}
\hline
Model         & \#Blocks   & Channel Dims            & Conv Type   \\ \hline
WaveDH        & [1, 2, 3]  & [32, 64, 128, 64, 32]    & GroupConv   \\ 
WaveDH-Tiny   & [1, 2, 2]  & [24, 48, 96, 48, 24]   & DWConv      \\ \hline
\end{tabular}}
\label{table1:config}
\end{table}

\begin{table*}[t]
\centering
\caption{Ablation study on the proposed components of WaveDH. We test the PSNR and SSIM results \\ on SOTS-indoor set, and compute the parameter numbers and MACs.}
\scalebox{1.1}{
\begin{tabular}{l|llll}
\hline
Models                  & PSNR  &SSIM    & \#Params (M) & \#MACs (G) \\
\hline
\rowcolor{skyblue}
\textbf{WaveDH (Full Model)}   & \textbf{39.35} &\textbf{0.995}  & 1.490       & 7.824      \\
w/o FU                  & 39.05 {\tiny \textcolor{red}{(-0.30)}} &0.994 {\tiny \textcolor{red}{(-0.001)}}               & 1.451 {\tiny \textcolor{blue}{(-0.039)}}       & 6.677 {\tiny \textcolor{blue}{(-1.147)}}     \\
w/o WA                  & 39.09 {\tiny \textcolor{red}{(-0.26)}} &0.994 {\tiny \textcolor{red}{(-0.001)}}     & 1.478 {\tiny \textcolor{blue}{(-0.012)}}      & 7.743 {\tiny \textcolor{blue}{(-0.081)}}     \\
w/o FU \& DU            & 39.08 {\tiny \textcolor{red}{(-0.27)}} &0.994 {\tiny \textcolor{red}{(-0.001)}}               & 1.513 {\tiny \textcolor{red}{(+0.023)}}      & 7.080 {\tiny \textcolor{blue}{(-0.744)}}     \\
w/o FU \& WA            & 38.91 {\tiny \textcolor{red}{(-0.44)}} &0.994 {\tiny \textcolor{red}{(-0.001)}}               & \textbf{1.439} {\tiny \textcolor{blue}{(-0.051)}}       & \textbf{6.596} {\tiny \textcolor{blue}{(-1.228)}}      \\
w/o FU \& WA \& DU      & 38.92 {\tiny \textcolor{red}{(-0.43)}} &0.994 {\tiny \textcolor{red}{(-0.001)}}               & 1.501 {\tiny \textcolor{red}{(+0.011)}}      & 6.999 {\tiny \textcolor{blue}{(-0.825)}}     \\
\hline
\end{tabular}}
\label{table2:ablation}
\end{table*}

\section{Experiments}
\label{sec:exp}

\subsection{Datasets and implementation}

\subsubsection{Datasets}
For our experiments, we select the RESIDE dataset \cite{RESIDE}, a widely acknowledged benchmark for single image dehazing. RESIDE is comprehensive, comprising five subsets: Indoor Training Set (ITS), Outdoor Training Set (OTS), Synthetic Objective Testing Set (SOTS), Real World task-driven Testing Set (RTTS), and Hybrid Subjective Testing Set (HSTS). Following the FFA-Net \cite{ffanet}, we use ITS (13,990 image pairs) and OTS (313,950 image pairs) for training WaveDH, testing performance on the indoor (500 image pairs) and outdoor (500 image pairs) sets of SOTS. Additionally, the I-HAZE \cite{ihaze} dataset, containing 30 image pairs of hazy and haze-free indoor scenes, is incorporated to diversify test scenarios. We evaluate the performance using Peak Signal-to-Noise Ratio (PSNR) and Structural Similarity Index (SSIM) to ensure a comprehensive comparison with state-of-the-art methods. The choice of these datasets and metrics is pivotal in illustrating our WaveDH's effectiveness across diverse hazing conditions.

\subsubsection{Implementation details}
The models are trained and tested separately for indoor and outdoor scenes. During the training stage, we train our models on ITS for 700 epochs and on OTS for 60 in RGB channels. In each training mini-batch, we randomly crop 32 patches of size $256 \times 256$ from hazy images as the input. The proposed model is optimized by minimizing L1 loss and the contrastive loss \cite{aecr} with AdamW optimizer \cite{adamw}. The initial learning rate is set to 0.0005 on ITS and 0.0002 on OTS, and the cosine annealing strategy \cite{cosine} is used to adjust the learning rate. All experiments are conducted with the PyTorch framework.

We present two models according to the number of feature channels and the type of convolution in FMBConv. The detailed configurations are reported in Table \ref{table1:config}. The training code and our models will be available on public.

 \begin{figure}[b]
 \centering
    \includegraphics[scale=0.26]{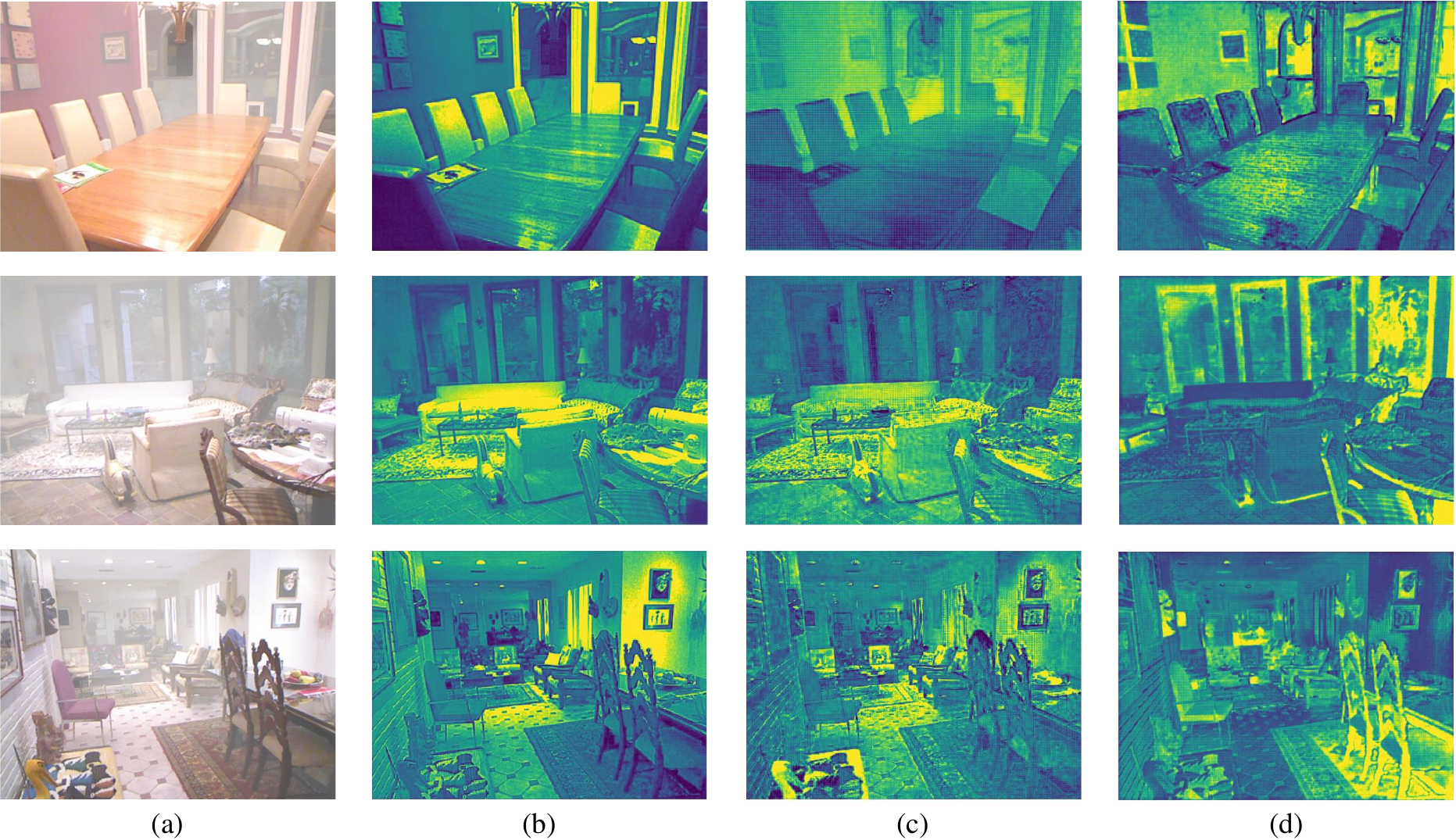}
    \caption{Feature map visualization of the dual-upsample and fusion mechanism. (a) Original hazy input images. (b) Feature maps after the Inverse Discrete Wavelet Transform (IDWT) layer. (c) Feature maps after the PixelShuffle layer. (d) Feature maps after fusion module. This highlights the effectiveness of the dual-upsample and fusion mechanism.}
    \label{fig5:DUandFU}
        \vspace{-0.5cm}
 \end{figure}

 \begin{figure*}[t]
 \centering
    \includegraphics[scale=0.58]{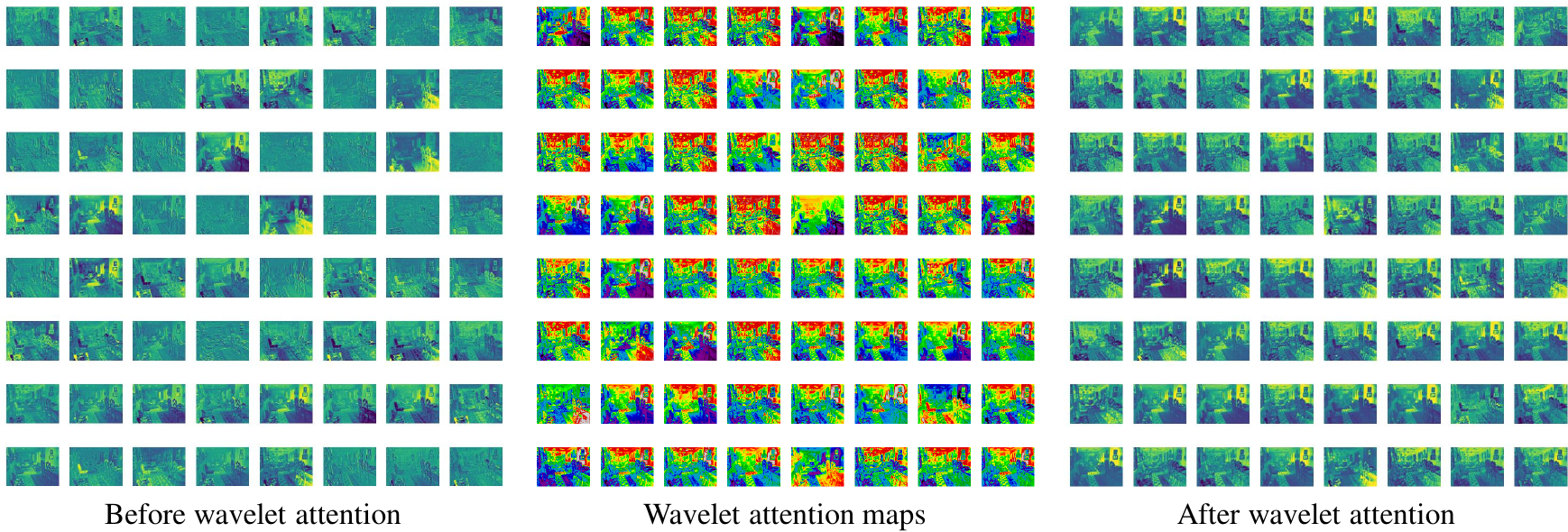}
    \caption{Comparative visualization of feature maps. This shows three sets of feature maps: before wavelet attention (\textit{left}), wavelet attention maps (\textit{center}), and after wavelet attention (\textit{right}). The transformation illustrates the enhancement of feature maps clarity and noise suppression achieved through the wavelet attention mechanism. Please zoom in for better visualization and best viewed on the screen.}
    \label{fig3:activation}
        \vspace{-0.10cm}
 \end{figure*}

\begin{table}[t]
\centering
\caption{Ablation study of the convolution type in FMBConv. We test the PSNR and SSIM results on SOTS-indoor set, and compute the parameter numbers and MACs.}
\scalebox{0.9}{
\begin{tabular}{l|llll}
\hline
Conv type              & PSNR & SSIM    & \#Params (M)   & \#MACs (G) \\ \hline
\rowcolor{skyblue}
\textbf{Group Conv}    & 39.35 & \textbf{0.995}  & 1.490          & 7.824       \\
Standard Conv          & \textbf{39.61} {\tiny \textcolor{blue}{(+0.26)}} & \textbf{0.995} {\tiny \textcolor{blue}{(+0.000)}}  & 2.729 {\tiny \textcolor{red}{(+1.239)}}          & 10.576 {\tiny \textcolor{red}{(+2.752)}}      \\
Depthwise Conv         & 38.72 {\tiny \textcolor{red}{(-0.63)}} & 0.994 {\tiny \textcolor{red}{(-0.001)}}  & \textbf{1.252} {\tiny \textcolor{blue}{(-0.238)}}         & \textbf{6.600} {\tiny \textcolor{blue}{(-1.224)}}      \\ \hline
\end{tabular}}
\label{table3:ablation}
\end{table}

\subsection{Ablation Studies}
\label{ablation}
In our comprehensive ablation experiments conducted on the SOTS-indoor dataset, we delve into the individual and collective contributions of three pivotal components of the WaveDH model: the Wavelet Attention (WA) module, the dual-upsampling (DU) mechanism, and the fusion (FU) module. Our in-depth analysis, summarized in Table \ref{table2:ablation}, quantifies the impact of each module on the dehazing performance. In addition to these core components, we also present an analysis of the convolution types within the Feature Mixing Block Convolution (FMBConv), detailed in Table \ref{table3:ablation}. Moreover, our study extends to examine the influence of the Group Convolution Expansion Ratio ($r_{conv}$) and the MLP Expansion Ratio ($r_{mlp}$) on the overall efficacy and efficiency of WaveDH, with findings presented in Tables \ref{table3:conv_ratio} and \ref{table3:mlp_ratio}.

\subsubsection{Analysis of dual-upsample and fusion mechanism}
Our ablation study begins with a detailed evaluation of the dual-upsample and fusion mechanism, which is strategically formulated to address the lack of high-frequency information. The dual-upsampling (DU) scheme comprises a pair of upsampling modules, Inverse Discrete Wavelet Transform (IDWT) and PixelShuffle. Our fusion module (FU) is designed to effectively fuse the channel-wise features from the dual-upsampling (DU) module, ensuring a detail-rich output. (See Fig. \ref{fig5:DUandFU})

To investigate the contribution of fusion module, we simplify the model by removing it and replacing the concatenation operation with an element-wise addition, a naive form of fusion. The experimental results presented in Table \ref{table2:ablation} indicate that the full WaveDH model achieves a PSNR of 39.35. When the fusion module is removed (w/o FU), there is a noticeable decrease in PSNR to 39.05, indicating a decrease of 0.30. Moreover, we evaluate the effectiveness of DU module in w/o FU model variant by excluding the IDWT layer and using only the PixelShuffle layer with $1 \times 1$ convolution for upsampling. Although this model variant (w/o FU \& DU) shows a slight increase in PSNR, it also resulted in increased computational costs and number of parameters, as indicated in Table \ref{table2:ablation}. These results demonstrate the effectiveness of the dual-upsample and fusion mechanism in enhancing dehazing performance while maintaining computational efficiency.

\begin{table}[t]
\centering
\caption{Ablation study on different group convolution expansion ratio $r_{conv}$  of FMBConv. We test the PSNR and SSIM results on SOTS-indoor set, and compute the parameter numbers and MACs.
}
\scalebox{0.9}{
\begin{tabular}{l|llll}
\hline
Ratio              & PSNR & SSIM    & \#Params (M)   & \#MACs (G) \\ \hline
\rowcolor{skyblue}
$r_{conv}$ = 1.5    & \textbf{39.35} & \textbf{0.995}  & 1.490          & 7.824       \\
$r_{conv}$ = 1.25   & 39.15 {\tiny \textcolor{red}{(-0.20)}} & \textbf{0.995} {\tiny \textcolor{blue}{(+0.000)}}  & \textbf{1.429} {\tiny \textcolor{blue}{(-0.061)}}          & \textbf{7.657} {\tiny \textcolor{blue}{(-0.167)}}      \\
$r_{conv}$ = 2     & 38.65 {\tiny \textcolor{red}{(-0.70)}} & 0.994 {\tiny \textcolor{red}{(-0.001)}}  & 1.612 {\tiny \textcolor{red}{(+0.122)}}         & 8.160 {\tiny \textcolor{red}{(+0.336)}}      \\ \hline
\end{tabular}}
\label{table3:conv_ratio}
\end{table}

\begin{table}[t]
\centering
\caption{Ablation study on different MLP expansion ratio $r_{mlp}$ of channel projection modules. We test the PSNR and SSIM results on SOTS-indoor set, and compute the parameter numbers and MACs.
}
\scalebox{0.9}{
\begin{tabular}{l|llll}
\hline
Ratio              & PSNR & SSIM    & \#Params (M)   & \#MACs (G) \\ \hline
\rowcolor{skyblue}
$r_{mlp}$ = 1.25    & 39.35 & \textbf{0.995}  & \textbf{1.490}          & \textbf{7.824}       \\
$r_{mlp}$ = 1.5     & \textbf{39.36} {\tiny \textcolor{blue}{(+0.01)}} & 0.994 {\tiny \textcolor{red}{(-0.001)}}  & 1.519 {\tiny \textcolor{red}{(+0.029)}}      & 7.875 {\tiny \textcolor{red}{(+0.051)}}      \\
$r_{mlp}$ = 2         & 38.55 {\tiny \textcolor{red}{(-0.80)}} & 0.994 {\tiny \textcolor{red}{(-0.001)}}  & 1.577 {\tiny \textcolor{red}{(+0.087)}}         & 7.975 {\tiny \textcolor{red}{(+0.151)}}      \\ \hline
\end{tabular}}
\label{table3:mlp_ratio}
\end{table}

\begin{table*}[t]
\centering
\caption{Quantitative comparison of various dehazing methods on SOTS-indoor and SOTS-outdoor in terms of PSNR, SSIM. We also report number of parameters (\#Params), number of floating-point operations (\#FLOPs) to perform comprehensive comparisons. The sign ``-'' indicates the digit is unavailable. Note that FLOPs are measured on 256 $\times$ 256 size images. \textcolor{red}{Red} represents the best performance while \textcolor{blue}{Blue} and \textcolor{green}{Green} indicate the second and third best performance, respectively.}
\scalebox{1.0}{
\begin{tabular}{c|cc|cc|cc}
\hline \multirow{3}{*}{Methods} & \multicolumn{2}{c|}{ITS} & \multicolumn{2}{c|}{OTS} & \multicolumn{2}{c}{\multirow{2}{*}{Overhead}} \\ \cline{2-5}
& \multicolumn{2}{c|}{SOTS-indoor} & \multicolumn{2}{c|}{SOTS-outdoor} \\ \cline{2-7} 
                        & PSNR  & SSIM  & PSNR  & SSIM   & \#Params (M) & \#MACs (G) \\
\hline 
(TPAMI’10) DCP \cite{DCP}                   & 16.62 & 0.818 & 19.13 & 0.815  & -       & -       \\
(TIP’16) DehazeNet \cite{dehazenet}                & 21.14 & 0.847 & 22.46 & 0.8514 & 0.009   & 0.514   \\
(ICCV’17) AOD-Net \cite{aodnet}                & 19.06 & 0.850 & 20.29 & 0.8765 & 0.0018  & 0.114   \\
(CVPR’18) GFN \cite{GFN}                    & 22.30 & 0.880 & 21.55 & 0.844  & 0.499   & 14.94   \\
(WACV’19) GCANet \cite{GCANet}                 & 30.23 & 0.980 & -     & -      & 0.702   & 18.41   \\
(ICCV’19) GDN \cite{Griddehazenet}                    & 32.16 & 0.984 & 30.86 & 0.982  & 0.956   & 18.77   \\
(CVPR’20) MSBDN \cite{MSBDN}                  & 33.67 & 0.985 & 33.48 & 0.982  & 31.35   & 41.54   \\
(ECCV’20) PFDN  \cite{pfdn}                  & 32.68 & 0.976 & -     & -      & 11.27   & 50.46   \\
(AAAI’20) FFA-Net \cite{ffanet}                & 36.39 & 0.989 & 33.57 & \textcolor{green}{0.984}  & 4.456   & 287.53  \\
(CVPR’21) AECR-Net \cite{aecr}                & 37.17 & 0.990 & -     & -      & 2.611   & 52.20   \\
(TIP’22) SGID-PFF \cite{sgid-pff}                 & 38.52 & 0.991 & 30.20 & 0.975  & 13.87   & 152.8   \\ 
(CVPR’22) MAXIM-2S \cite{maxim}               & 38.11 & 0.991 & 34.19 & \textcolor{blue}{0.985}  & 14.1    & 216     \\
(AAAI’22) UDN \cite{udn}                    & \textcolor{green}{38.62} & 0.991 & \textcolor{red}{34.92} & \textcolor{red}{0.987}  & 4.25    & -       \\
(TIP'23) DehazeFormer-M (\cite{dehazeformer}) & 38.46 & \textcolor{blue}{0.994} & 34.29 & 0.983 & 4.634 & 48.64 \\
(TIP'24) DEA-Net-S (\cite{chen2024dea}) & \textcolor{blue}{39.16} & \textcolor{green}{0.992} & - & - & 2.844 & 24.88 \\
\hline 
\rowcolor{skyblue}
\textbf{WaveDH-Tiny}    & 36.93     & \textcolor{green}{0.992}     & \textcolor{green}{34.52}     & 0.983      & 0.543        & 3.507       \\
\rowcolor{skyblue}
\textbf{WaveDH}         & \textcolor{red}{39.35} & \textcolor{red}{0.995} & \textcolor{blue}{34.89}     & \textcolor{green}{0.984}      & 1.490    & 7.824   \\
\hline
\end{tabular}}
\label{table3:comparison}
\end{table*}

\subsubsection{Analysis of wavelet attention (WA) module}
We also study the impact of the Wavelet Attention (WA) module, which plays a pivotal role in selectively suppressing noise while simultaneously enhancing the low-frequency context and the high-frequency detail information critical for effective dehazing. To quantify the importance of WA, we conduct experiments on a model variant devoid of the WaveAttention module (w/o WA), simplifying the block to only include Discrete Wavelet Transform (DWT) and a $1 \times 1$ convolution layer. The results, as depicted in Table \ref{table2:ablation}, reveal that removing WA from WaveDH leads to a substantial decrease in PSNR by 0.26, from 39.35 to 39.09, highlighting the efficacy of the WA module in our dehazing process. In terms of computational complexity, the inclusion of WA leads to a slight increase in the number of parameters and Multiply-Accumulates (MACs) by 0.012 million and 0.081 billion, respectively. This minimal rise in complexity reflects the efficiency of the WA module in enhancing the model capability without heavily burdening computational resources.

Furthermore, we visualize the feature maps before and after applying wavelet attention to provide a clearer perspective, as shown in Fig. \ref{fig3:activation}. The enhanced feature maps with wavelet attention reveal distinctly clearer backgrounds and objects by effectively suppressing noisy pixels. It is evident that the WA module yields a more informative representation due to the rich contour features present in the high-frequency components, which are adeptly captured by our WA module that leverages not only low-frequency component but also high-frequency components.

\begin{figure*}[t]
  \centering
  \includegraphics[scale=0.56]{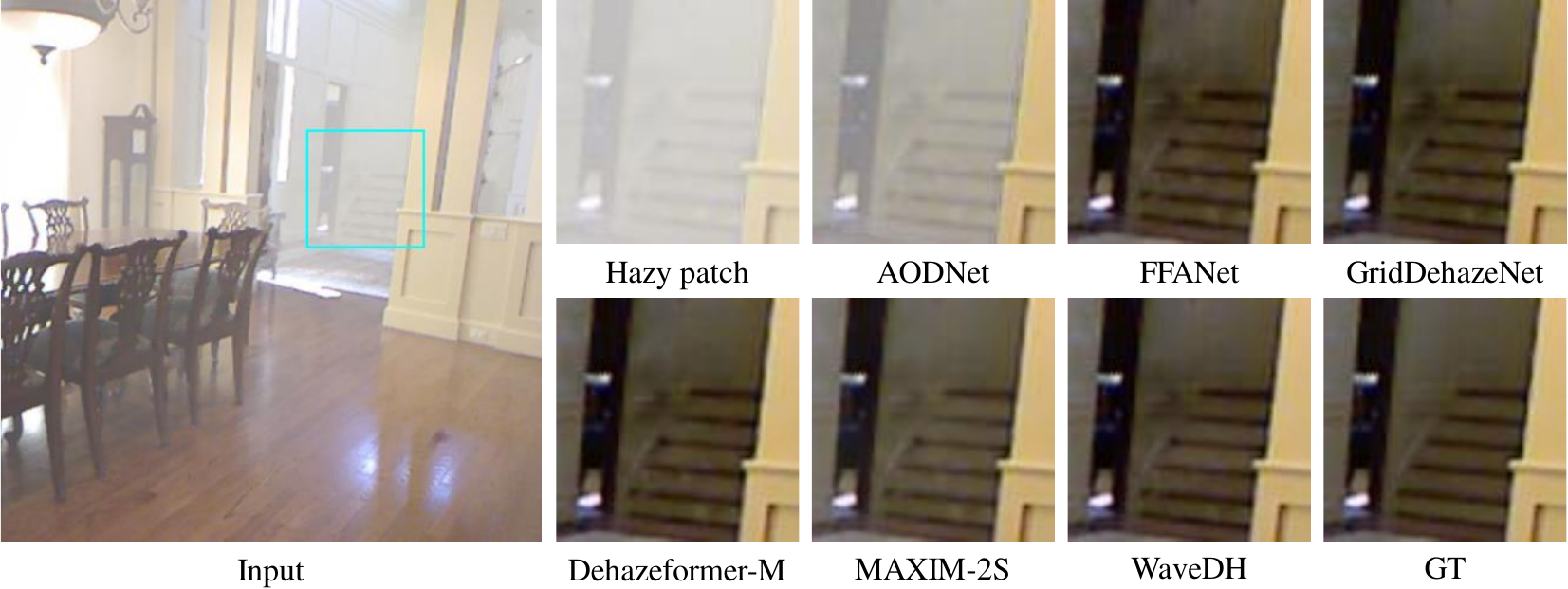}\par
  \vspace{1em}
  \includegraphics[scale=0.56]{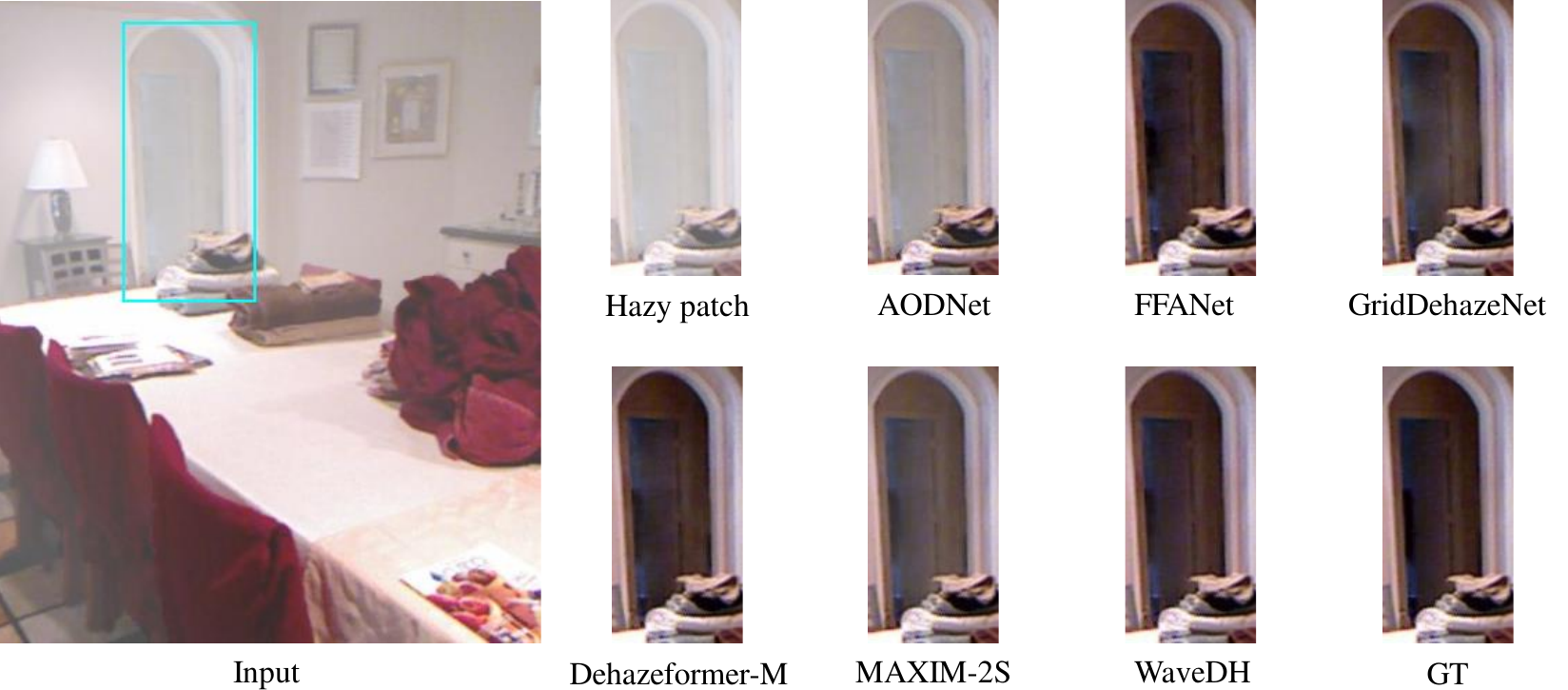}\par
  \caption{Qualitative comparison with the state-of-the-art methods on SOTS-indoor dataset.}
  \label{fig4:indoor}
\end{figure*}

\subsubsection{Analysis of Convolution Type in FMBConv}
In this part of our ablation study, we focus on evaluating the impact of different convolution types used in the Fused-MBConv (FMBConv) on the performance of WaveDH. We compare three convolution types: Standard Convolution, Group Convolution, and Depthwise Convolution. The results of this study are summarized in Table \ref{table3:ablation}. \textit{Standard Convolution}: Achieves the highest PSNR of 39.61 and a SSIM of 0.995. However, it also resulted in the highest number of parameters (2.729 million) and the greatest computational cost, as indicated by 10.576 billion MACs. \textit{Group Convolution}: Demonstrates a slightly lower PSNR of 39.35 and an SSIM of 0.995. It significantly reduces the number of parameters to 1.490 million and the computational cost to 7.824 billion MACs, presenting a more balanced trade-off between performance and efficiency. \textit{Depthwise Convolution}: Shows the lowest PSNR of 38.72. It further reduces the number of parameters to 1.252 million and the computational cost to 6.600 billion MACs, indicating its efficiency in terms of resource utilization. 

These results emphasize the trade-offs between convolution types in terms of dehazing performance, parameter efficiency, and computational cost. After a careful consideration of these results, we opt for group convolution in our WaveDH model. This decision is anchored on a well-considered trade-off between performance and computational efficiency. While standard convolution offers slightly better performance, its high computational demand and parameter size are less suited for scenarios where resources are limited or efficiency is paramount. On the other hand, depthwise convolution, despite its impressive efficiency, falls short in maintaining the dehazing performance that our model aims to achieve.

\begin{table}[t]
\centering
\caption{Quantitative comparison of various dehazing methods on the I-Haze in terms of PSNR, SSIM. \textcolor{red}{Red} and \textcolor{blue}{Blue} indicate the second and third best performance, respectively.}
\scalebox{1.0}{
\begin{tabular}{c|ccc}
\hline
Methods              & PSNR        & SSIM       & \#MACs (G)   \\ \hline
AOD-Net \cite{aodnet}             & 14.74       & 0.669      & 0.114    \\
GCANet \cite{GCANet}              & 15.64       & 0.709      & 18.41    \\
GDN \cite{Griddehazenet}                 & 15.12       & 0.710      & 18.77    \\
FFANet \cite{ffanet}              & 15.57       & 0.701      & 287.53       \\
DehazeFlow \cite{dehazeflow}          & 15.28       & 0.695      & -       \\
MAXIM-2S \cite{maxim}            & \textcolor{red}{16.25}       & \textcolor{blue}{0.718}      & 216      \\
D4 \cite{d4}                   & 15.61       & 0.702      & 2.246       \\
\hline
\rowcolor{skyblue}
\textbf{WaveDH-Tiny} & \textcolor{blue}{15.91}       & \textcolor{red}{0.720}      & 3.507     \\
\rowcolor{skyblue}
\textbf{WaveDH}      & \textcolor{blue}{15.91}       & 0.708      & 7.824      \\ \hline
\end{tabular}}
\label{table4:ihaze}
\end{table}

\begin{figure*}[t]
  \centering
  \includegraphics[scale=0.56]{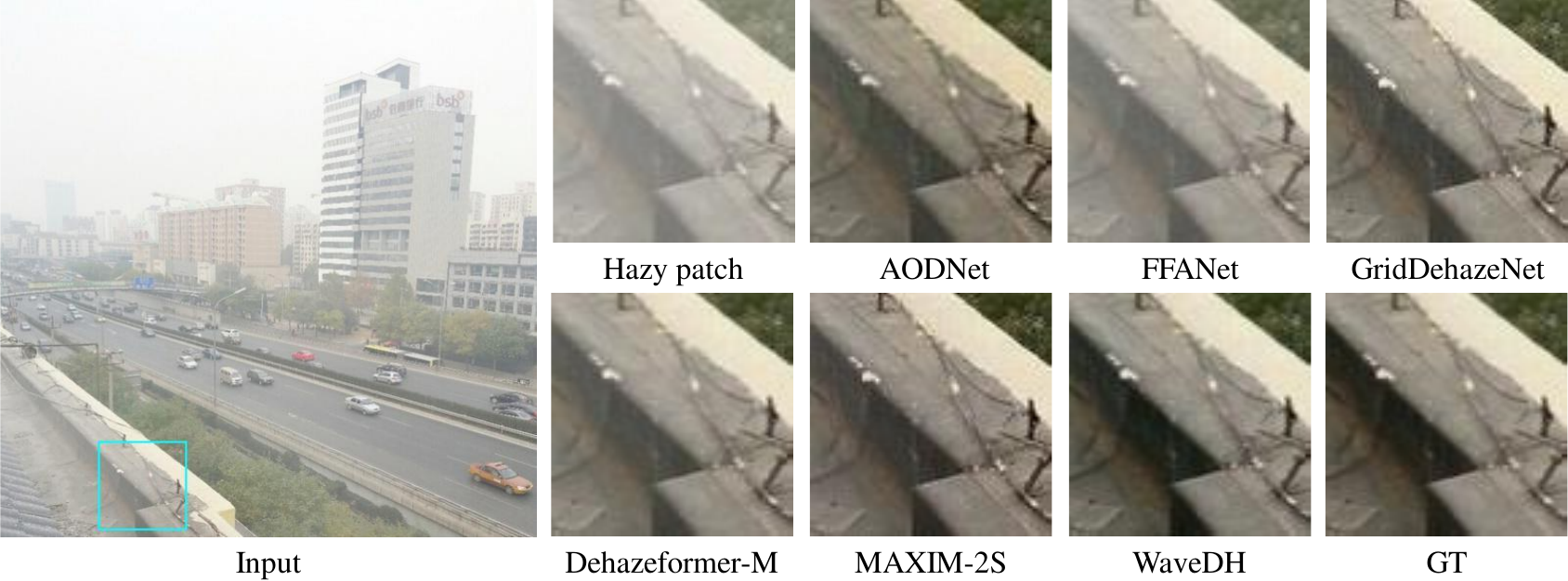}\par
  \vspace{1em}
  \includegraphics[scale=0.56]{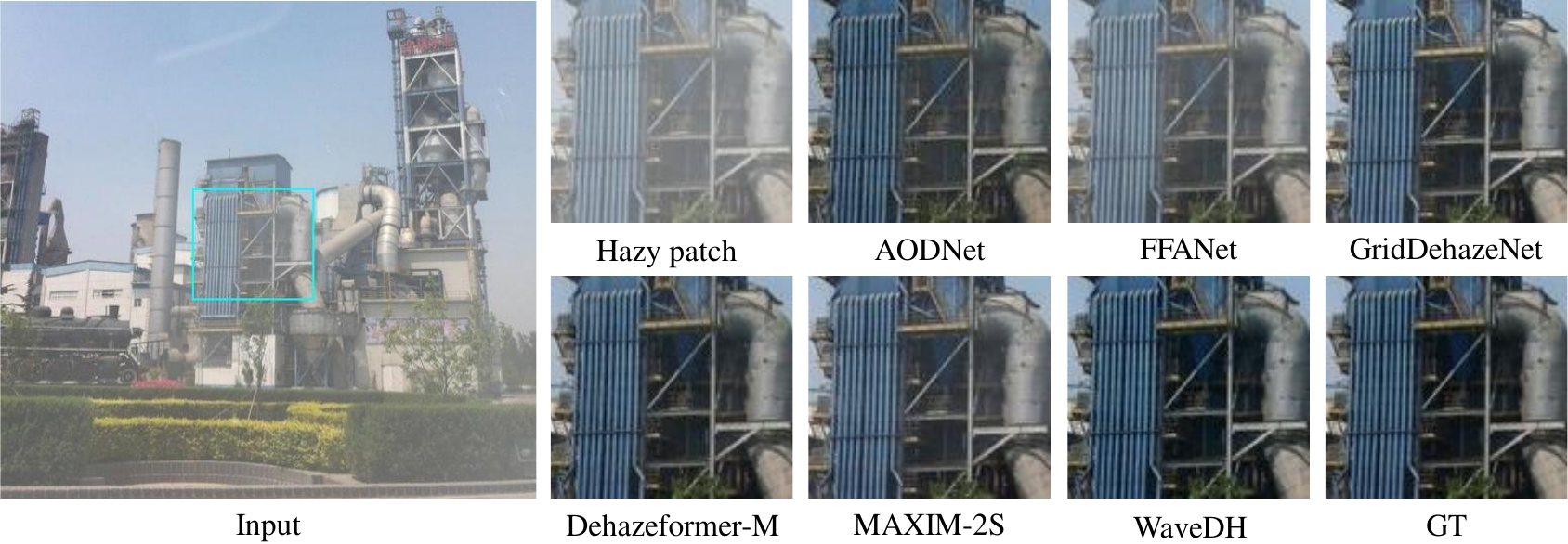}\par
  \caption{Qualitative comparison of with state-of-the-art methods on SOTS-outdoor dataset.}
  \label{fig5:outdoor}
\end{figure*}

\begin{figure*}[t]
  \centering
  \includegraphics[scale=0.492]{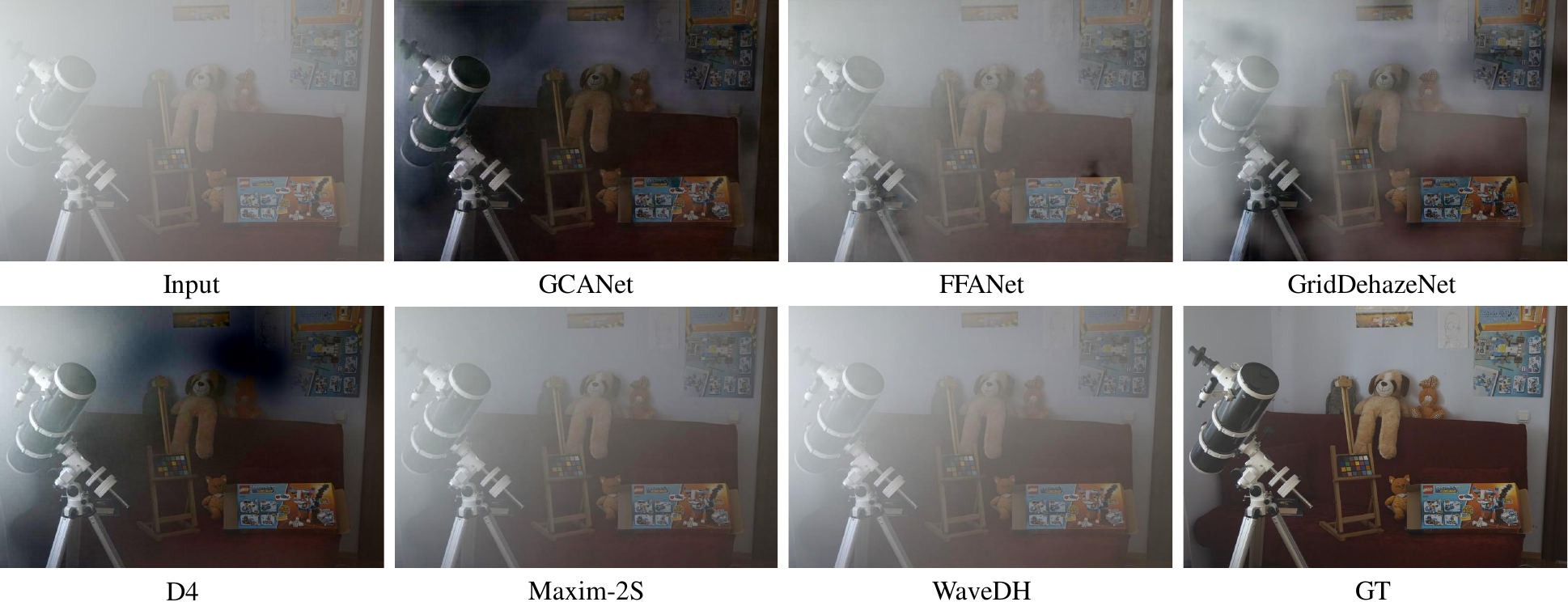}\par
  \vspace{1em}
  \includegraphics[scale=0.492]{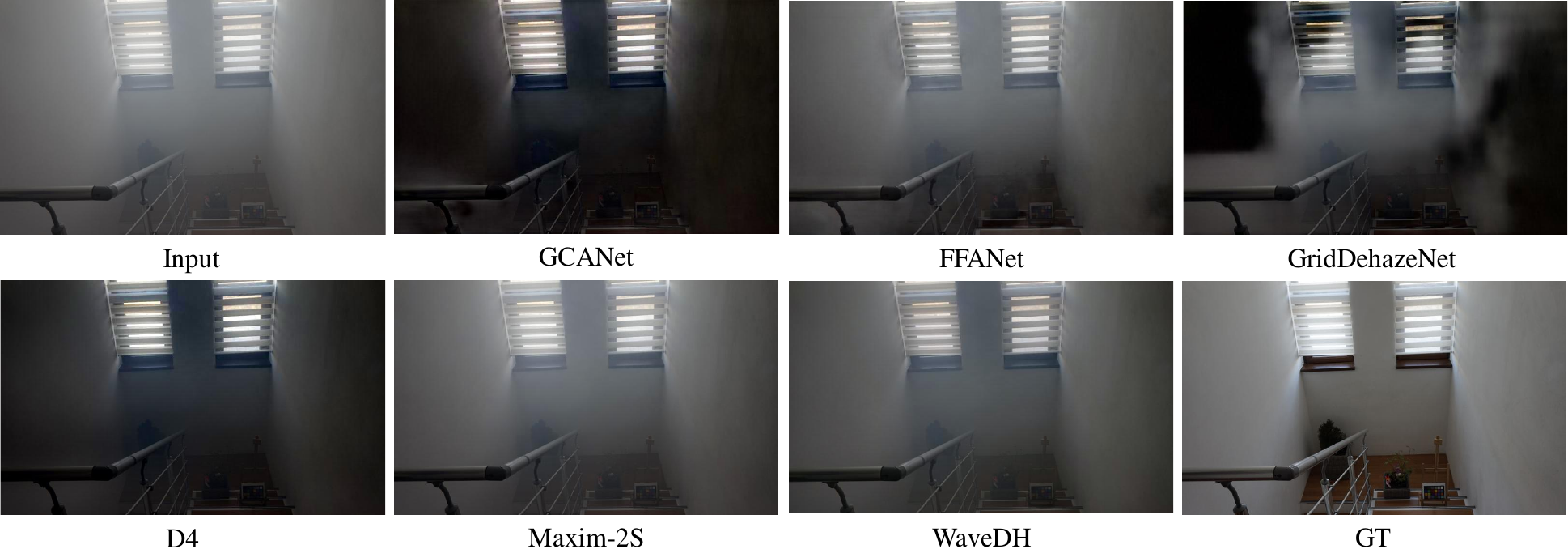}
  \caption{Qualitative comparison with the state-of-the-art methods on I-HAZE dataset.}
  \label{fig6:ihaze}
\end{figure*}

\subsubsection{Analysis of Group Convolution Expansion Ratio in FMBConv}
A critical hyperparameter of the Fused-MBConv (FMBConv) in our proposed WaveDH is the group convolution expansion ratio, denoted as $r_{conv}$, which has a direct bearing on the model capacity and efficiency. We evaluate the influence of $r_{conv}$ on image dehazing performance through quantitative analysis.

Establishing $r_{conv}$ at 1.5, our WaveDH achieves an optimal balance between dehazing quality and computational efficiency. This baseline configuration yields a PSNR of 39.35 and an SSIM of 0.995, with a manageable number of parameters of 1.490 million and 7.824 billion MACs, ensuring a balanced computational cost and performance relationship.

Reducing $r_{conv}$ to 1.25 leads to a slightly more efficient model, as reflected by the reduced number of parameters of 1.429 million and computational expense of 7.657 billion MACs. Nonetheless, this change incurs a discernible performance degradation with a PSNR drop to 39.15. On the other hand, increasing $r_{conv}$ to 2 leads to a marked decrease in PSNR, falling to 38.65, signaling a counterproductive trade-off with an increase in expansion leading to decreased performance. This outcome demonstrates the drawbacks of an aggressive expansion strategy that fails to yield commensurate improvements in dehazing quality. Based on these observations, we conclude that an expansion ratio of 1.5 within FMBConv is pivotal for optimizing the image dehazing performance of WaveDH.

\subsubsection{Analysis of MLP Expansion Ratio in channel projection module}
In the pursuit of fine-tuning the Feature Mixing Block (FMB), we investigate the influence of varying MLP expansion ratios ($r_{mlp}$) on the network dehazing efficacy and computational efficiency. The MLP expansion ratio controls the channel expansion within the pointwise convolutional layers. This ablation analysis studies how different values of $r_{mlp}$ affect the PSNR, SSIM, number of parameters, and MACs on the SOTS-Indoor dataset.

For baseline setting ($r_{mlp} = 1.25$), the model achieves a PSNR of 39.35 and an SSIM of 0.995, demonstrating high fidelity in dehazed images while maintaining structural details. This configuration also recorded the lowest computational footprint with 1.490 million parameters and 7.824 billion MACs, indicating a high level of efficiency.

Increasing $r_{mlp}$ to 1.5, the model exhibits a marginal PSNR improvement of 0.01, achieving a PSNR of 39.36 and a minor decrease in SSIM (-0.001) However, this slight increase in PSNR performance comes at the cost of an increase in the number of parameters (+0.029 million) and MACs (+0.051 billion). Despite these increases, the balance between dehazing quality and computational demand remains acceptable.

A more pronounced expansion with $r_{mlp}$ set to 2 results in a significant decrease in PSNR, from 39.35 to 38.55, indicating potential over-parameterization of channel projection module, which may not contribute positively to the dehazing capability. This suggests that a high MLP expansion ratio might introduce redundancy and inefficiencies without a proportional gain in dehazing quality. Our analysis clearly shows that an $r_{mlp}$ of 1.25 is the most plausible choice for our WaveDH, striking an optimal balance between performance enhancement and computational efficiency.

\subsection{Comparison With SOTA Methods}
\label{com_sota}
In this section, we provide both quantitative and qualitative comparison to analyze our novel WaveDH approach with existing state-of-the-art (SOTA) dehazing methods based on a diverse set of synthetic and real-world hazy images. For a fair comparison, we use the pretrained models provided by the authors.

\subsubsection{Quantitative Comparison}
Our quantitative evaluation leverages the Synthetic Objective Testing Set (SOTS), including both indoor and outdoor hazy image datasets, as the benchmark for performance metrics including PSNR and SSIM. As outlined in Table \ref{table3:comparison}, our WaveDH not only exhibits a remarkable reduction in parameters and computational complexity but also surpasses SOTA methods in performance metrics. Specifically, within the indoor dataset, our WaveDH achieves best results, recording a PSNR of 39.35dB and an SSIM of 0.995, outshining the DEA-Net-S by 0.19 dB in PSNR while requiring only 31\% of computational cost. For the SOTS-outdoor dataset, our WaveDH-Tiny model, with just 543K parameters, ranks third in PSNR, surpassing many existing methods. This demonstrates the strong capability of our WaveDH to handle haze on synthetic dataset.

For a real-world comparison, we use the I-HAZE dataset \cite{ihaze}, which, notably, was not used during the training phase, ensuring an unbiased evaluation. Here, both WaveDH and the WaveDH-Tiny models demonstrate competitive performance, achieving the second-highest PSNR values as shown in Table \ref{table4:ihaze}. Most impressively, WaveDH-Tiny leads in SSIM metrics, indicating the strengths of our methods when it comes to restoring clear scenes from real-world hazy observations.

\subsubsection{Qualitative Comparison}
We extended our evaluation to a qualitative perspective, selecting representative samples for a visual analysis to juxtapose the dehazing performance of WaveDH with other approaches. Figs \ref{fig4:indoor}, \ref{fig5:outdoor}, and \ref{fig6:ihaze} serve as visual comparisons on SOTS-indoor, SOTS-outdoor, and I-HAZE datasets, respectively.

Within the SOTS-indoor dataset, we analyze the network performance to dehaze two distinct scenes. As illustrated in Fig. \ref{fig4:indoor}, AOD-Net fails to address the haze, leaving much of it unmitigated. Both FFANet and GridDehazeNet, while attempting to dehaze, introduce severe color distortions that result in outputs that are excessively bright or dim. DehazeFormer-M and MAXIM-2S, though they somewhat mitigate color distortion issues, still appears to persist in some areas and introduce artifacts. In contrast, our WaveDH exhibits a commendable restoration of color fidelity across all image regions, preserving finer details and exhibiting minimal artifacts. The qualitative comparison of the SOTS-outdoor dataset, as demonstrated in Fig. \ref{fig5:outdoor}, indicates that most compared methods fail to remove the haze effectively. On the other hand, our WaveDH stands out by successfully reconstructing hazy-free scenes with preserved textural and color information, with the least haze residual.

Moving to real-world scenarios with the I-HAZE dataset, the complexity increases as the sample distribution significantly deviates from the training data. This shift challenges most methods, causing them to struggle with dehazing, as seen in Fig. \ref{fig6:ihaze}. The methods, including GCANet, GridDehazeNet, FFANet, and D4, are prone to severe color distortion and the production of artifacts to varying extents. However, MAXIM-2S and our WaveDH distinguish themselves by generating visually convincing results that are virtually artifact-free, reinforcing the superior generalizability of WaveDH. When considering both the visual results and the efficiency of the model, our approach, WaveDH, emerges as a particularly compelling solution for single image dehazing.

\section{Conclusion}
In this article, we proposed WaveDH, a novel ConvNet designed for efficient single-image dehazing. Central to our approach is the strategic use of wavelet sub-bands for guided up-and-downsampling, coupled with a novel frequency-aware feature refinement process. The novel squeeze-and-attention scheme in the downsampling block and the dual-upsample and fusion mechanism in the upsampling block are especially noteworthy, enhancing high-frequency detail reconstruction while optimizing computational costs. Our feature refinement block refines the intermediate features through a coarse-to-fine strategy, thereby enhancing the model efficiency and contributing to a well-calibrated balance between accuracy and computational cost. Comprehensive experiments validate our WaveDH's superior performance over existing state-of-the-art methods, achieving high-quality dehazing with reduced computational demands. However, our method  has limitations in real-world hazy scenes, particularly in recovering dense haze regions. Therefore, in our future research, we plan to address this issue and extend our approach to other low-level vision tasks.

\bibliographystyle{IEEETran}
\bibliography{bibliography}

\end{document}